%% file: main.tex
\pdfoutput=1

\documentclass[11pt]{article}

\usepackage[final]{acl}

\usepackage{times}
\usepackage{latexsym}

\usepackage[T1]{fontenc}

\usepackage[utf8]{inputenc}

\usepackage{microtype}

\usepackage{inconsolata}

\usepackage{graphicx}

\usepackage{bm}
\usepackage{amssymb}
\usepackage{amsmath}
\usepackage{enumitem}
\usepackage{tcolorbox}
\usepackage{caption,subfigure,graphics,epstopdf,wrapfig}
\usepackage{threeparttable,tabularx,booktabs,multirow,multicol,verbatim}
\usepackage{lipsum}
\usepackage{listings}

\definecolor{dark_red}{RGB}{189, 74, 87}
\makeatletter
\newcommand{\smallsym}[2]{#1{\mathpalette\make@small@sym{#2}}}
\newcommand{\make@small@sym}[2]{%
  \vcenter{\hbox{$\m@th\downgrade@style#1#2$}}%
}
\newcommand{\downgrade@style}[1]{%
  \ifx#1\displaystyle\scriptstyle\else
    \ifx#1\textstyle\scriptstyle\else
      \scriptscriptstyle
  \fi\fi
}
\makeatother
\newcommand{\sparagraph}[1]{\noindent \textbf{#1}}
\newcommand{\smallsim}{\smallsym{\mathrel}{\sim}}

\title{DiffLM: Controllable Synthetic Data Generation via Diffusion Language Models}

\author{Ying Zhou$^{1,2}$\thanks{Equal contribution.}, Xinyao Wang$^{3*}$\thanks{Work done before joining Amazon AGI.}, Yulei Niu$^{3\ddagger}$, Yaojie Shen$^{1,2}$, Lexin Tang$^{3}$, Fan Chen$^{3}$, \\
{\bfseries Ben He$^{1,2\ddagger}$, Le Sun$^{2}$, Longyin Wen$^{3}$\thanks{Corresponding authors.}} \\
\textsuperscript{1}University of Chinese Academy of Sciences, Beijing, China \\
\textsuperscript{2}Chinese Information Processing Laboratory, Institute of Software, \\
Chinese Academy of Sciences, Beijing, China \\
\textsuperscript{3}ByteDance Inc., San Jose, USA \\
{\tt zhouying20@mails.ucas.ac.cn}
}

\begin{document}
\maketitle
\begin{abstract}
Recent advancements in large language models (LLMs) have significantly enhanced their knowledge and generative capabilities, leading to a surge of interest in leveraging LLMs for high-quality data synthesis. 
However, synthetic data generation via prompting LLMs remains challenging due to LLMs' limited understanding of target data distributions and the complexity of prompt engineering, especially for structured formatted data. 
To address these issues, we introduce DiffLM, a controllable data synthesis framework based on variational autoencoder (VAE), which further (1) leverages diffusion models to reserve more information of original distribution and format structure in the learned latent distribution and (2) decouples the learning of target distribution knowledge from the LLM's generative objectives via a plug-and-play latent feature injection module. 
As we observed significant discrepancies between the VAE's latent representations and the real data distribution, the latent diffusion module is introduced into our framework to learn a fully expressive latent distribution.
Evaluations on seven real-world datasets with structured formatted data (i.e., Tabular, Code, and Tool data) demonstrate that DiffLM generates high-quality data, with performance on downstream tasks surpassing that of real data by 2\%–7\% in certain cases. 
Data and code are available at \url{https://github.com/bytedance/DiffLM}.
\end{abstract}

\input{tex/introduction}
\input{tex/realted_works}
\input{tex/methodology}
\input{tex/experiments}
\input{tex/analysis}
\input{tex/conclusion}

\section*{Limitations}
Our work, while offering valuable contributions to synthetic data generation, has several limitations. First, our approach focuses primarily on structured data synthesis, such as tabular data generation, code generation, and tool generation. It does not address the generation of more subjective or nuanced data, such as sentiment-specific data or tasks requiring varied levels of complexity or capability. Additionally, our decoupling of data distribution learning may need further refinement to accommodate such subjective text synthesis tasks.
Second, due to time and hardware constraints, we limited our study to a relatively simple VAE and diffusion network, and used LLMs up to 7B parameters for decoding. Future research could explore the potential of scaling these models further, using even larger architectures to better evaluate and validate the effectiveness of the DiffLM framework in data synthesis tasks.

\section*{Acknowledgments}
This work is supported by the National Natural Science Foundation of China (62272439) and the Fundamental Research Funds for the Central Universities, and is conducted during Ying Zhou's internship at ByteDance Inc.

\bibliography{main_ref}

\clearpage
\appendix
\input{tex/appendix}

\end{document}

%% file: tex/introduction.tex
\section{Introduction}

Data Synthesis has become an indispensable technique in current machine learning research, enabling rapid generation and modification of datasets~\citep{DBLP:journals/corr/abs-2401-02524}, allowing researchers to experiment with various scenarios and model architectures without the extensive processes associated with real-world data collection. 
Meanwhile, with the rapid advancements in large language models (LLMs), recent research in natural language processing (NLP) has increasingly focused on leveraging LLMs for synthetic data generation. 
Early efforts attempted to fine-tune LLMs to align with real data distributions~\citep{DBLP:journals/corr/abs-1909-05858,DBLP:conf/aaai/Anaby-TavorCGKK20,DBLP:conf/iclr/BorisovSLPK23}. 
As the in-context learning capabilities of LLMs have improved, some studies have explored zero-shot or few-shot prompting of LLMs to generate synthetic data~\citep{DBLP:conf/emnlp/YeGLXF00K22,DBLP:journals/corr/abs-2403-18802}.

Despite the progress achieved, generating high-quality synthetic textual data using LLMs remains challenging, particularly for structured data~\citep{DBLP:conf/emnlp/JosifoskiSP023,DBLP:conf/nips/LiTGLH22}. 
First, LLMs often lack a global understanding of the target data distribution when generating synthetic data. Even after fine-tuning, it is difficult to inject information about complex and varied distributions into current LLM architectures, often resulting in outputs with low diversity and instances of data copying~\citep{DBLP:journals/corr/abs-2406-18966,DBLP:conf/nips/YuZZMRKSZ23}. 
Moreover, existing LLM-based synthetic data generation methods typically involve complex pipelines and post-processing mechanisms, such as prompt engineering, multi-agent frameworks, and iterative sampling~\citep{DBLP:conf/iclr/Dekoninck0BV24,DBLP:journals/corr/abs-2406-18966}. These complexities hinder the rapid adaptation of LLMs to new tasks, limiting their utility in dynamic research and industrial scenario.
Concurrently, the remarkable performance of variational autoencoders (VAEs) and diffusion models in image synthesis tasks~\citep{betker2023dalle,DBLP:conf/cvpr/RombachBLEO22} has spurred interest in adapting these techniques to other modalities~\citep{DBLP:conf/iclr/BorisovSLPK23,DBLP:conf/nips/LiTGLH22,DBLP:conf/iclr/GongLF0K23}. 
Although some works have introduced latent spaces into language models for simple tasks like style transfer or topic generation~\citep{DBLP:conf/naacl/YangK21,DBLP:conf/nips/LiTGLH22}, our preliminary experiments indicate that directly applying the latent distributions learned by VAEs often results in outputs that are unrelated to the real data. 
Similar issues also have been addressed in prior works~\citep{DBLP:journals/corr/abs-2402-10575,DBLP:journals/corr/abs-2004-14758,DBLP:conf/conll/BowmanVVDJB16}. This challenges the direct application of these methods in more complex scenarios for synthetic data generation.

To address these challenges, we propose DiffLM, a novel framework that leverages a plug-and-play latent space to provide data distribution information for LLMs during data generation. 
First, to decouple the learning of real data distributions from the LLM’s training objectives, we develop a latent space using a VAE model to capture external information, mapping samples from the real dataset to latent vectors. 
However, we observed that sampling points from a Gaussian distribution obtained from naive VAE that cannot generate realistic results. To overcome the poor quality of data generated by sampling from VAE, we employ a latent diffusion method that linearly adds noise to the latent space over time. A denoising network is then trained to learn these noises in the reverse process, reducing efficiency loss in data synthesis due to sampling failures. 
Finally, we design a soft prompting method to inject latent features into the LLM decoding process, resulting in controllable, high-quality synthetic data.
We evaluate our method on seven real-world structured formatted datasets, ranging from relatively simple table synthesis to more complex code and tool synthesis tasks. Experiments demonstrate that DiffLM can generate high-quality results, and ablation studies confirm the effectiveness of each component in our proposed method. 
The contributions of this paper are threefold:
\begin{itemize}[leftmargin=*, itemsep=-0.2em, topsep=0.2em]
\item {\bfseries Decoupling Data Distribution Learning}: We proposed a new VAE-based LLM framework for data systhesis, which decouples the learning of real data distribution information from the training objectives of the LLM by introducing the a small projection network.
\item {\bfseries High-Quality Synthetic Data}: Based on our observations, the meticulously designed VAE and diffusion structures effectively model the distribution of real data, enabling the generation of high-quality synthetic data. In all tasks, the quality of the generated data is comparable to or even surpasses that of the real data.
\item {\bfseries Comprehensive Evaluation}: We validate the high quality of data generated by DiffLM across three distinct scenarios and seven datasets, underscoring its robustness and adaptability in advancing synthetic data generation for natural language processing.
\end{itemize}

%% file: tex/realted_works.tex
\section{Related Works}

\paragraph{Large Language Models for Data Synthesis.}
The recent advancement in the generative capabilities of LLMs has motivated numerous exploratory works aiming to leverage these models for data augmentation in areas such as text classification~\citep{DBLP:conf/emnlp/YeGLXF00K22,DBLP:conf/emnlp/LiZL023}, information extraction~\citep{DBLP:journals/corr/abs-2303-04360,DBLP:conf/emnlp/JosifoskiSP023}, and tabular data generation~\citep{DBLP:conf/iclr/BorisovSLPK23,DBLP:journals/corr/abs-2406-14541}. 
A comprehensive survey conducted by \citet{DBLP:conf/acl/LongWXZDCW24} proposes a prompt-based generic workflow for synthetic data generation, curation, and evaluation. 
And multiple advanced works have attempted to fine-tune language models for data synthesis in recent years~\citep{DBLP:conf/aaai/Anaby-TavorCGKK20,DBLP:journals/corr/abs-2003-02245,DBLP:conf/nips/DinhZZLGRSP022,DBLP:conf/iclr/BorisovSLPK23,DBLP:journals/corr/abs-2406-14541}. Specifically, these methods involve fine-tuning LLMs on a small amount of gold data for language modeling, followed by the use of various sampling methods to generate data.
However, a major challenge remains in ensuring that synthetic data accurately reflects real-world distributions. \citet{DBLP:journals/corr/abs-2305-15041} has shown that LLM-generated data can sometimes diverge from actual data distributions, leading to unfaithful representations that may hinder model training.
Some studies have explored data selection~\citep{DBLP:conf/emnlp/PuriSSPC20} or data augmentation~\citep{DBLP:conf/emnlp/YeG0F0K22} to address this distribution gap, but there remains significant room for improvement.

\input{figure/framework}

\paragraph{Latent Variable Models for Text Generation.}
Latent variable models have made significant advances in computer vision in recent years~\citep{DBLP:journals/tmlr/YuXKLBWVKYAHHPLZBW22,DBLP:conf/cvpr/GuCBWZCYG22,DBLP:journals/corr/abs-2310-04378,DBLP:conf/iclr/GulrajaniKATVVC17}, achieving high-quality generation results, flexibility and effectiveness, as well as robustness to noise perturbations.
In particular, latent diffusion models, such as DALL-E~\citep{betker2023dalle} and Stable Diffusion~\citep{DBLP:conf/cvpr/RombachBLEO22}, operate their diffusion processes in a latent space rather than directly in data space, enabling a near-optimal balance between generation quality and computational efficiency. 
In text generation, several works~\citep{DBLP:conf/conll/BowmanVVDJB16,DBLP:journals/corr/abs-1801-09797,DBLP:journals/corr/abs-2004-14758,DBLP:conf/nips/LiTGLH22,DBLP:conf/acl/GuFMZGZ023,DBLP:conf/iclr/BorisovSLPK23,DBLP:journals/corr/abs-2402-10575} have attempted to combine latent spaces with language models to accomplish tasks such as 
language modeling~\citep{DBLP:conf/icml/LinGSWFLDC23,DBLP:conf/iclr/ReidHN23,DBLP:conf/acl/0011LZ23,DBLP:conf/naacl/YuanYTHH24}, unconditional text generation~\citep{DBLP:journals/corr/abs-2206-05895,DBLP:conf/emnlp/ChenZ0SY23} and control text generation~\citep{DBLP:conf/nips/AustinJHTB21,DBLP:conf/acl/HeSTWHQ23}.
Additionally, some studies have explored the use of diffusion for plug-and-play controllable generation~\citep{DBLP:conf/nips/LiTGLH22,DBLP:conf/iclr/GongLF0K23}, aiming to steer the outputs of pre-trained language model using auxiliary modules. 
While these works share a similar perspective with ours, we tackle a more challenging scenario of structured data synthesis.
To the best of our knowledge, our work is the first to combine VAEs and denoising diffusion models with large language models for high-quality data synthesis.


%% file: figure/framework.tex
\begin{figure*}[t]
\centering
\includegraphics[width=\textwidth]{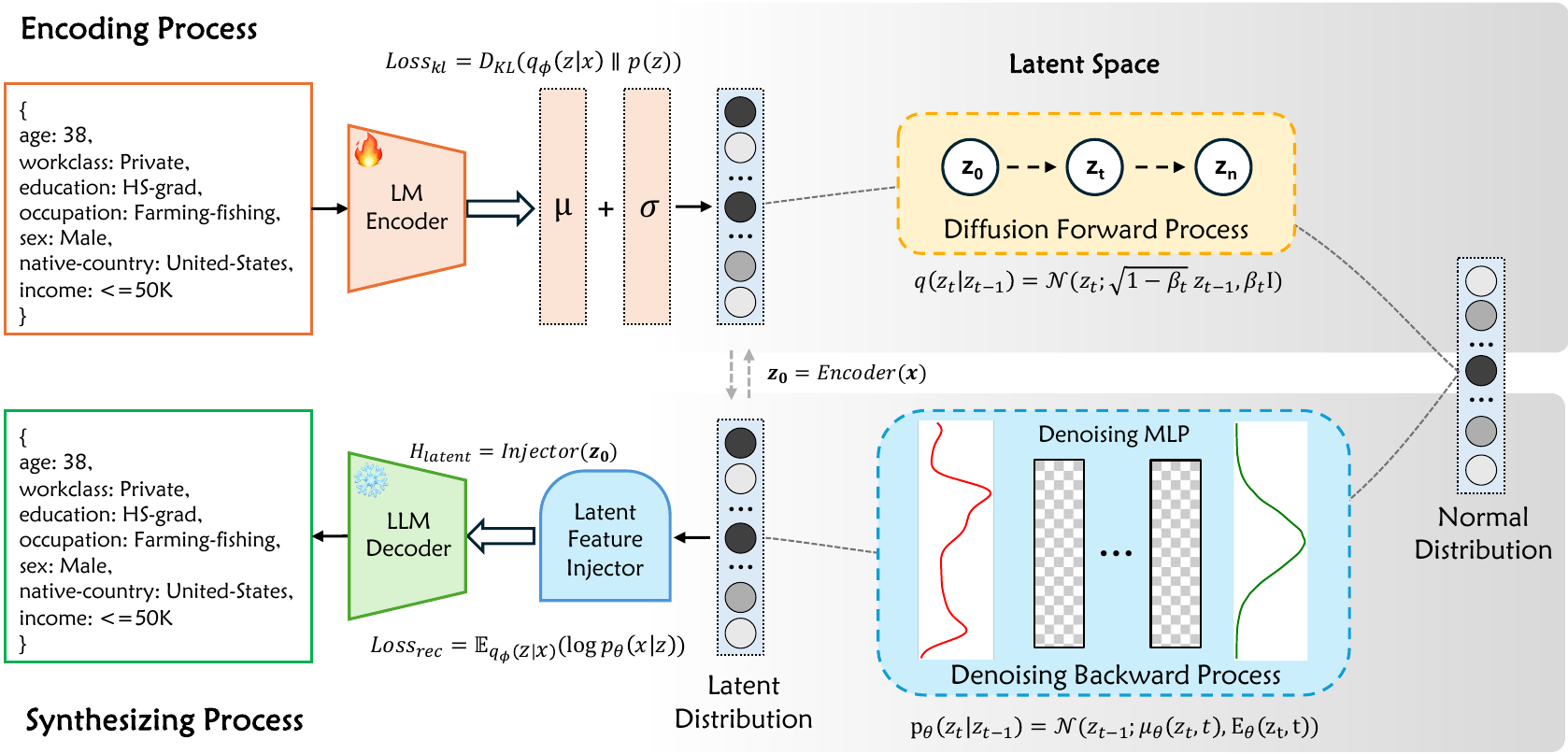}
\caption{Overview of our \textbf{DiffLM}. The trainable lanaguage model (LM) works as VAE encoder while the fixed LLM decoder serves as VAE decoder. We further (1) introduced a Diffusion module to learn the latent space, and (2) employ a latent feature injector with soft prompting to align latent vector space with LLM decoder.}
\label{fig:framework}
\end{figure*}

%% file: tex/methodology.tex
\section{Methodology}
Figure~\ref{fig:framework} illustrates the main pipeline of our proposed DiffLM. 
First, we define an encoder to map discrete text into a continuous latent space (Section~\ref{sec:method:vae}). 
Second, although the features of the data are extracted and compressed, conventional latent embeddings in text VAEs often lead to decoding failures due to underutilized or empty regions in the latent space. To address this issue, we train a diffusion model on the latent space (Section~\ref{sec:method:diff}). 
Finally, to incorporate encoded prior knowledge into the decoding stage of large language models, we propose a novel soft prompt injection method to steer the decoding process (Section~\ref{sec:method:injection}).

\subsection{Problem Formulation}\label{sec:method:definition}

We begin by defining $\mathcal{D}$ as a known small set of real-world distribution data, where each element $x$ represents a real sample.
We define $G$ as the synthetic data generator, which learns the distribution of $\mathcal{D}$ and generates a set of synthetic samples, $\mathcal{D}_{syn}$, ensuring that the model does not simply memorize and reproduce the same real samples, meaning $\mathcal{D}\cap \mathcal{D}_{syn} = \varnothing$. 
It should be noted that we focus on the task of unconditional data synthesising using LLMs, where $G$ generates synthetic samples independently of any additional context, i.e., without using explicit prompt text.

\subsection{VAE-based Representation Learning}\label{sec:method:vae}

\sparagraph{Feature Encoding:} In standard VAEs, an encoder is typically employed to map input data into a latent space.
Given structured text data $s_i$, we utilize a learnable Transformer-based pre-trained language model~\citep{DBLP:conf/nips/VaswaniSPUJGKP17,DBLP:conf/naacl/DevlinCLT19,DBLP:journals/jmlr/RaffelSRLNMZLL20} to obtain the representation vector $x_i \in \mathbb{R}^{d \times 2}$, which can be split into the mean and variance. 
Using the re-parameterization trick~\citep{DBLP:journals/corr/KingmaW13}, we then obtain the latent feature $z \in \mathbb{R}^{d}$:
\begin{equation}
z = \mu + \sigma \odot \epsilon,
\end{equation}
where $\mu$ and $\sigma$ are the mean and standard deviation output by the encoder, and $\epsilon$ is sampled from a standard normal distribution $\mathcal{N}(0, I)$.

\sparagraph{LLM Decoding:} After generating the latent feature $z$, we employ a frozen-parameter LLM to reconstruct the input text $s$ in a causal language modeling manner.
The rationale for freezing the LLM parameters is to avoid retraining and to preserve its general knowledge and reasoning capabilities.
Consequently, aligning the two different modalities, whereas the latent space and the LLM input space, presents a significant challenge. 
To address this, we propose a novel latent feature injector using soft prompting and design a corresponding injector network; specific details are provided in Section~\ref{sec:method:injection}.

\sparagraph{VAE Model Training Objective:} The VAE model is typically trained using the Evidence Lower Bound (ELBO) loss function.
Following previous work~\citep{DBLP:journals/corr/abs-1804-03599}, we adopt the $\beta$-VAE training strategy~\citep{DBLP:conf/iclr/HigginsMPBGBML17}, which introduces a weighting parameter $\beta$ to control the contribution of the KL divergence loss in the total loss function. 
Specifically, when $\beta = 0$, the model reduces to a standard autoencoder. For $\beta > 0$, the KL constraint encourages learning a smoother latent space:
\begin{align}
\text{ELBO}_{\beta} &= L_{rec} - \beta L_{kl}, \\
L_{rec} &= \mathbb{E}_{q_\phi(z|x)}\big(\log p_\theta(x|z)\big), \\
L_{kl} &= D_{\text{KL}}\big(q_\phi(z|x)~\|~p(z)\big),
\end{align}
where $p_\theta(x|z)$ is the language modeling reconstruction likelihood, $q_\phi(z|x)$ is the approximate posterior, and $p(z)$ is the prior over the latent space, i.e., Gaussian distribution.
In our model design, considering the denoising network of latent diffusion, we adopt an decreasing $\beta$ adjustment strategy. 
We initially set a larger $\beta$ weight to enforce a strong regularization on the latent space. 
As the reconstruction loss convergence slows, we decrease the $\beta$ value to allow the model to focus more on reconstruction accuracy. 
Additionally, we employ an early stopping mechanism to prevent overfitting.

\subsection{Latent Space Denoising}\label{sec:method:diff}
Although VAE can learns latent space representations of data, directly sampling from the prior distribution $p(z)$ often exhibit low quality generated samples.
In our preliminary experiments, we observed that directly utilizing the latent features learned by the VAE frequently produces text that is unrelated to the target data distribution.
This issue arises due to the discrepancy between the encoder's learned posterior distribution $q_\phi(z|x)$ and the prior $p(z)$. 
To address this problem, we introduce a diffusion model in the latent space to more accurately model the true distribution of the latent features.
Inspired by \citet{DBLP:conf/iclr/Zhang0SSQFRK24}, we extract the latent vectors $z \in \mathcal{Z}$ from the trained VAE for each data point $x \in \mathcal{D}_{\text{train}}$. 
Starting from the initial latent vector $z_0$, we progressively add noise over time following a linear schedule to get $z_t$.
During the reverse diffusion process, we employ a standard continuous denoising network to recover $z_0$~\citep{DBLP:conf/iclr/0011SKKEP21}.
For the training objective, we optimize the diffusion model through denoising score matching~\citep{DBLP:conf/nips/KarrasAAL22}:
\begin{equation}
z_t = z_{0} + \sigma(t)\epsilon, \epsilon \in \mathcal{N}(0, I), \label{eq:diff_forward}
\end{equation}
\begin{equation}
\begin{aligned}
dz_t = &-\dot{\sigma}(t)\sigma(t)\nabla_{z_t} \log{p(z_t)} dt \\ &+ \sqrt{2\dot{\sigma}(t)\sigma(t)} d\omega_t. \label{eq:diff_backward}
\end{aligned}
\end{equation}
In forward process Eq.\ref{eq:diff_forward}, $z_t$ is the latent variable at time $t$,
and $\sigma(t)$ is a time-dependent noise scale function.
As for backward process Eq.\ref{eq:diff_backward},
$\dot{\sigma}(t)$ stands for the time derivative of $\sigma(t)$,
and $\nabla_{z_t} \log{p(z_t)}$ is the gradient of the log probability density with respect to \( z_t \), also known as the score function, and $d\omega_t$ is an increment of the Wiener process (standard Brownian motion).
As for diffusion model training loss Eq.\ref{eq:diff_loss},
$\epsilon_\theta(z_t, t)$ is the neural network that predicts the noise $\epsilon$ given $z_t$ and $t$.
\begin{equation}
\mathcal{L}_{\text{diff}} = \mathbb{E}_{t \smallsim p(t),~z_0\smallsim 
p(z_0),~\epsilon \smallsim \mathcal{N}(0, I)} \left\| \epsilon_\theta(z_t, t) - \epsilon \right\|^2, \label{eq:diff_loss}
\end{equation}
The detailed description for diffusion model could be found in Appendix~\ref{sec:app:diffusion}.

\input{table/performance_tabular}

\subsection{Latent Feature Injection}\label{sec:method:injection}
After constructing a latent space that captures the true data distribution, two challenges remain: 
1) 
\textit{Aligning latent space with LLM's input space}.
How can the decoding LLM process the latent vector $z$ to steer a powerful language model for realistic data generation?
2) \textit{Seamless Integration with LLM Knowledge}: How can we integrate external information without disrupting the LLM's internal knowledge?
Motivated by adapter training methods in LLM fine-tuning~\citep{DBLP:conf/emnlp/LesterAC21,DBLP:conf/acl/LiL20,DBLP:conf/icml/HoulsbyGJMLGAG19,DBLP:conf/nips/LiuLWL23a}, we consider the soft prompt latent injection approach to incorporate $z$ into LLM decoding without training the model weights. 
Specifically, after obtaining the latent representation $z$, we use an upper MLP to map it into $k$ soft prompt token embeddings, denoted as $\mathbf{H_{\text{latent}}} \in \mathbb{R}^{k \times d}$. These soft embeddings serve as a steering vector, which is concatenated before the $<$BOS$>$ token to assist the LLM in decoding. The detailed process is illustrated in Figure~\ref{fig:injections}.
We also conduct ablation experiments in Section~\ref{sec:ablation} with the other two injection methods proposed by \citet{DBLP:conf/emnlp/LiGLPLZG20}, which validated that our methods obtain the best reconstruction loss and downstream task performance.






%% file: table/performance_tabular.tex
\begin{table*}[t!]
\caption{Performance of downstream tasks using generated \textbf{tabular} data. 
We evaluate the quality from: performance in machine learning efficiency (\textbf{MLE}) task, and column-wise distribution density estimation ($\bm \rho$) task. $\uparrow, \downarrow$ indicate that higher (or lower) metrics correspond to better performance.
\textbf{Boldface} indicates DiffLM surpasses the SoTA model based on language models. \textcolor{dark_red}{\bfseries Red Boldface} denotes DiffLM exceeds the performance achieved using real data.
}
\begin{center}
\resizebox{\linewidth}{!}{%
\begin{tabular}{ @{} l  *{10}{c} @{} }
\toprule[1.0pt]
\multirow{2}{*}{\bfseries Method} &  \multicolumn{2}{c}{\bfseries Adult} & \multicolumn{2}{c}{\bfseries Default} & \multicolumn{2}{c}{\bfseries Magic} & \multicolumn{2}{c}{\bfseries Shoppers} & \multicolumn{2}{c}{\bfseries Beijing} \\
\cmidrule(lr){2-3} \cmidrule(lr){4-5} \cmidrule(lr){6-7}  \cmidrule(lr){8-9}  \cmidrule(lr){10-11}
& {MLE~$\uparrow$} & {$\rho$~$\downarrow$} & {MLE~$\uparrow$} & {$\rho$~$\downarrow$} & {MLE~$\uparrow$} & {$\rho$~$\downarrow$} & {MLE~$\uparrow$} & {$\rho$~$\downarrow$} & {MLE~$\downarrow$} & {$\rho$~$\downarrow$} \\
\midrule
Real & 0.927 & - & 0.770 & - & 0.946 & - & 0.926 & - & 0.423 & - \\
SMOTE & 0.899 & 1.60 & 0.741 & 1.48 & 0.934 & 0.91 & 0.911 & 2.68 & 0.593 & 1.85 \\
\midrule
CTGAN & 0.886 & 16.84 & 0.696 & 16.83 & 0.855 & 9.810 & 0.875 & 21.15 & 0.902 & 21.39 \\
TVAE & 0.878 & 14.22 & 0.724 & 10.17 & 0.887 & 8.250 & 0.871 & 24.51 & 0.770 & 19.16 \\
GOGGLE & 0.778 & 16.97 & 0.584 & 17.02 & 0.654 & 1.900 & 0.658 & 22.33 & 1.090 & 16.93 \\
CoDi & 0.871 & 21.38 & 0.525 & 15.77 & 0.932 & 11.56 & 0.865 & 31.84 & 0.818 & 16.94 \\
TabSyn & 0.915 & 0.58 & 0.764 & 0.85 & 0.938 & 0.88 & 0.920 & 1.43 & 0.582 & 1.12 \\
\midrule
GPT-4$_{ICL}$ & 0.889 & 32.55 & - & - & 0.864 & 10.07 & 0.835 & 41.11 & 0.992 & 26.1 \\
Mistral-7B$_{ICL}$ & 0.803 & 29.22 & 0.732 & 36.00 & 0.881 & 25.64 & 0.882 & 42.00 & 0.865 & 12.45 \\
Qwen2.5-14B$_{ICL}$ & 0.848 & 17.28 & 0.737 & 27.77 & 0.831 & 22.72 & 0.783 & 52.69 & 1.289 & 27.46 \\
GReaT & 0.913 & 12.12 & 0.755 & 19.94 & 0.888 & 16.16 & 0.902 & 14.51 & 0.653 & 8.25 \\
{DiffLM ({\itshape ours})} & 0.906 & {\bfseries 9.74} & \textcolor{dark_red}{\bfseries 0.794} & {\bfseries 9.06} & {\bfseries 0.917} & {\bfseries 7.53} & {\bfseries 0.915} & {\bfseries 10.07} & 0.696 & {\bfseries 6.35} \\
\bottomrule[1.0pt]
\end{tabular}
}
\end{center}
\label{tab:tabular}
\end{table*}

%% file: tex/experiments.tex
\section{Experiments}

In this section, we evaluate the generation quality of the DiffLM method on multiple public benchmarks across three tasks:
1) Tabular Data Generation: We compare DiffLM with SoTA tabular generation algorithms, demonstrating its strong capability in structured data generation. 
2) Code Generation: DiffLM showcases the ability to integrate structured data priors with its internal knowledge. The results on synthetic data are even better than real ones.
3) Tool Generation: DiffLM can quickly adapt to complex function call scenarios, highlighting its flexibility and adaptability.
All experiments adopt Mistral-v0.3 as the frozen VAE decoder. Additional implementation details for reproduction can be found in Appendix~\ref{sec:app:reproduction}.

\input{table/performance_code}

\subsection{Tabular Data Generation}

\sparagraph{Benchmarking.} We selected five publicly available datasets for evaluation, encompassing both classification and regression tasks: Adult, Beijing, Default, Magic, and Shoppers. The properties of
datasets are presented in Table~\ref{tab:app:tabular_dataset_deails}.
To assess the quality of synthetic data, we employed two perspectives: 1) \textbf{Low-order statistical metrics}, where we quantified column-wise density estimation using the Kolmogorov-Smirnov Test for numerical columns and the Total Variation Distance for categorical columns;
2) \textbf{Downstream task performance}, where we measured the predictive accuracy on test data of classifiers or regressors trained on the generated data. 

\sparagraph{Baselines.} We selected a comprehensive set of classic and SoTA tabular data generation models with diverse architectures for comparison. 
First, we consider the performance on real data as the upper bound for evaluation. 
Secondly, we included the classic method, synthetic minority over-sampling technique (SMOTE)~\citep{DBLP:journals/jair/ChawlaBHK02}, which generates new synthetic data patterns by performing linear interpolation between minority class samples and their $k$ nearest neighbors. 
Additionally, for neural network-based tabular generation algorithms, we considered six baselines across different architectures: 
1) \textit{GAN}-based models: CTGAN~\citep{DBLP:conf/nips/XuSCV19}; 
2) \textit{VAE}-based models: TVAE~\citep{DBLP:conf/nips/XuSCV19}, GOGGLE~\citep{DBLP:conf/iclr/LiuQBS23}; 
3) \textit{Diffusion}-based models: CoDi~\citep{DBLP:conf/icml/Lee0P23}, TabSyn~\citep{DBLP:conf/iclr/Zhang0SSQFRK24}; 
4) \textit{LLM}-based: 5-shot ICL-based generation, where structural information for each dataset (specifically, column names and data types for each row) is also provided. We leverage GPT-4, Mistral-v0.3-7B-instruct, and Qwen2.5-14B-instruct models for synthetic data generation. Additionally, we include GReaT~\citep{DBLP:conf/iclr/BorisovSLPK23}, which attempts to fine-tune a GPT-2~\citep{radford2019language} for table synthesis. We also report the GReaT result with Mistral model~\citep{DBLP:journals/corr/abs-2310-06825} in Appendix~\ref{sec:app:great_mistral}.
It is worth noting that we compare with the SoTA generative models not to merely outperform them in tabular generation but to demonstrate that our flexible DiffLM can achieve comparable performance while offering additional advantages.

\sparagraph{Evaluation.} Table~\ref{tab:tabular} presents the quality assessment results of the generated data. 
For different tabular datasets, we train an XGBoost classifier or regressor on the synthetic data to predict label values, evaluating performance using AUC and RMSE. 
The results indicate that the performance of the LLM-based synthetic data generation methods leveraging ICL is generally lower; even GPT-4, the best-performing ICL-based model, consistently achieves inferior results across all datasets compared to the fine-tuned GReaT model. However, our proposed DiffLM method consistently outperforms the current LLM-based SoTA GReaT on most datasets.
Notably, on the \textit{Default} dataset, the prediction accuracy using DiffLM’s synthetic data surpasses that obtained by training on real data. 
This suggests that DiffLM’s approach of integrating the real data distribution with its own learned knowledge can provide richer information for downstream tasks while preserving the original data structure.
In other words, the synthetic data generated by DiffLM contains additional knowledge compared to real data, which is challenging to achieve with previous methods.
Moreover, our generated results achieve performance comparable to prior methods in column-wise distribution density estimation. 
Although the TabSyn method attains superior performance on several datasets, it should be noted that our approach focuses on general, pluggable generation control for large language model, rather than training data synthesis models from scratch for specific domains. 
Despite this, in tabular data generation, our method’s performance is on par with these domain-specific methods.

\subsection{Code Generation}

\sparagraph{Benchmarking.} In the code generation scenario, to simplify the problem, we focus on Python code and use the Flytech\footnote{\url{https://huggingface.co/datasets/flytech/python-codes-25k}} dataset as real data, which contains 24,813 unique real user queries and the corresponding Python code fulfilling those requests. We discard the user queries and use only the code to train DiffLM. 
After generating synthetic code data, we continue pre-training the  Mistral 7B v0.3 base model~\citep{DBLP:journals/corr/abs-2310-06825} using a smaller learning rate, i.e., 1e-5, in a causal language modeling objective.
We then benchmark the trained model on code generation tasks, selecting two mainstream benchmarks: HumanEval~\citep{DBLP:journals/corr/abs-2107-03374} and MBPP~\citep{DBLP:journals/corr/abs-2108-07732}.
To better understand the performance changes of the base model, we also experiment larger models like Mistral Nemo with 12B parameters.

\sparagraph{Baselines.} We include the performance from recent code LLMs. 
First, we consider the CodeLLaMA~\citep{DBLP:journals/corr/abs-2308-12950} series, which use approximately 600B tokens to continue pre-training the LLaMA-2~\citep{DBLP:journals/corr/abs-2307-09288} base model, injecting strong code capabilities through multi-task learning.
Additionally, we compare with the Mistral base model~\citep{DBLP:journals/corr/abs-2310-06825} and its instruction-tuned variants, the latter could representing the upper bound of code capabilities for this architecture.

\sparagraph{Evaluation.} We report the code generation capabilities in Table~\ref{tab:code}. Specifically, Mistral-Real-Code and Mistral-DiffLM-Code denote models that were further pre-trained on real data and synthetic data generated by DiffLM, respectively. 
The 7B models are based on Mistral-0.3-Base, and the 12B models are based on Mistral-Nemo-Base. Both models were trained for 3 epochs on the same amount of data using identical hyperparameters, effectively serving as a controlled experiment where the data source is the only variable. 
The results indicate that simply continuing to pre-train the Mistral model with a small amount of code data leads to inconsistent impacts on code generation capabilities. 
Specifically, Mistral-Real-Code shows a slight improvement on \textit{HumanEval} but a significant decline on \textit{MBPP}. 
However, using our synthetic data to continue pre-training the base model yields better results than using real data. For instance, Mistral-DiffLM-Code-7B, achieved a 7\% improvement over the base model, even outperforming the Code Llama 7B model that was trained with more extensive data.
In summary, in the code generation scenario, we focus on the differing impacts of real data and synthetic data, further demonstrating that DiffLM can generate synthetic data that is even more effective than real data in enhancing downstream task performance.

\input{figure/tool_score}
\input{table/tool_win_rate}

\input{figure/eval_loss}

\subsection{Tool Generation}
\sparagraph{Evaluation.} To address more complex structured data generation scenarios, we further conduct a tool synthesis task. 
Specifically, we select the ToolBench~\citep{DBLP:conf/iclr/QinLYZYLLCTQZHT24} dataset as a benchmark for comparison, which is constructed based on the RapidAPI\footnote{\url{https://rapidapi.com/hub}} platform by crawling APIs created by real users and synthesizing related dialogue SFT data using GPT\footnote{\url{https://chat.openai.com}}. 
We use the its toolset to train DiffLM and then sample an equal number of tools for comparison.
We assess the usability of the generated tools from two perspectives:
1) {\bfseries Single-Tool Quality}: We use GPT-4 as an annotator to score the real and synthetic data across multiple dimensions on a scale from 0 to 10, where the results are illustrated in Figure~\ref{fig:gpt_score}.
2) {\bfseries Category-Level Preference}: We collect all tools within the same category and use GPT-4 to perform preference scoring between real tools and synthetic tools, as presented in Table~\ref{tab:tool_win_rate}. 
The specific evaluation prompts are provided in the appendix~\ref{sec:app:prompt}.
From the results, DiffLM’s synthetic data achieves higher scores in the single-tool scoring task, indicating that leveraging the internal knowledge and generative capabilities of LLMs allows us to create tool descriptions and input/output parameter definitions of higher textual quality. 
Additionally, in the category-level preference evaluation, nearly $1/3$ of the tool types surpass or are on par with real data in terms of diversity and usability. Since DiffLM can sample and generate tools indefinitely to increase coverage, we believe there is room for further improvement in this metric.

%% file: table/performance_code.tex
\begin{table*}[t!]
\centering
\caption{pass@k scores on \textbf{HumanEval} and \textbf{MBPP}. 
We follow \citet{DBLP:journals/corr/abs-2107-03374} for estimating pass@k, where $n>k$ solutions are generated per problem with p = 0.95 and a temperature of 0.2 to calculate the success rate with zero-shot learning.
{\bfseries Boldface} indicates that DiffLM surpasses the performance achieved with real data. 
\textcolor{dark_red}{\bfseries Red Boldface} indicates that DiffLM surpasses the base model’s performance.
}

\begin{threeparttable}
\begin{center}
\resizebox{0.95\linewidth}{!}{%
\begin{tabular}{ @{} ll | *{3}{c} | *{3}{c} @{} }
\toprule[1.0pt]
\multirow{2}{*}{{\bfseries Model}} & \multirow{2}{*}{{\bfseries Size}} & \multicolumn{3}{c}{\bfseries HumanEval} & \multicolumn{3}{c}{\bfseries MBPP} \\
\cmidrule(lr){3-5} \cmidrule(lr){6-8}
& & {pass@1} & {pass@10} & {pass@100} & {pass@1} & {pass@10} & {pass@100} \\
\midrule
GPT-4 & - & 67.00 & - & - & - & - & - \\
\multirow{2}{*}{CodeLLaMA} & 7B & 33.50 & 59.60 & 85.90 & 41.40$^*$ & 66.70$^*$ & 82.50$^*$ \\
& 34B & 48.80 & 76.80 & 93.00 & 55.00$^*$ & 76.20$^*$ & 86.60$^*$ \\
\midrule
\multirow{2}{*}{Mistral-Base} & 7B & 27.79 & 41.22 & 56.37 & 37.31 & 52.02 & 59.65 \\
& 12B & 10.12$^\dag$ & 20.91$^\dag$ & 28.93$^\dag$ & 43.38 & 61.44 & 69.09 \\\midrule
\multirow{2}{*}{Mistral-Instruct} & 7B & 36.09 & 52.95 & 64.18 & 38.45 & 50.77 & 59.17 \\
& 12B & 7.08$^\dag$ & 12.43$^\dag$ & 16.14$^\dag$ & 52.20 & 63.61 & 69.02 \\
\midrule
\multirow{2}{*}{\bfseries Mistral-Real-Code} & 7B & 28.58 & 42.24 & 54.24 & 27.15 & 42.21 & 48.14 \\
& 12B & 36.97 & 52.04 & 60.95 & 34.79 & 45.49 & 50.22 \\
\midrule
\multirow{2}{*}{\bfseries Mistral-DiffLM-Code} & 7B & \textcolor{dark_red}{\bfseries 35.37} & \textcolor{dark_red}{\bfseries 47.36} & {\bfseries 54.38} & {\bfseries 32.70} & 41.65 & 47.39 \\
& 12B & \textcolor{dark_red}{\bfseries 42.24} & \textcolor{dark_red}{\bfseries 56.02} & \textcolor{dark_red}{\bfseries 61.97} & \textcolor{dark_red}{\bfseries 44.42} & {\bfseries 52.35} & {\bfseries 55.70} \\
\bottomrule[1.0pt]
\end{tabular}
}
\end{center}

\small
\begin{tablenotes}
\item[*] These results are evaluated under a 3-shot setting.
\item[\dag] The vanilla Mistral-Nemo 12B models fail to pass the HumanEval benchmark, resulting in a lower score. We have conducted multiple evaluations and report the average performance.
\end{tablenotes}
\end{threeparttable}

\label{tab:code}
\end{table*}

%% file: figure/tool_score.tex
\begin{figure}[t]
\centering
\includegraphics[width=0.48\textwidth]{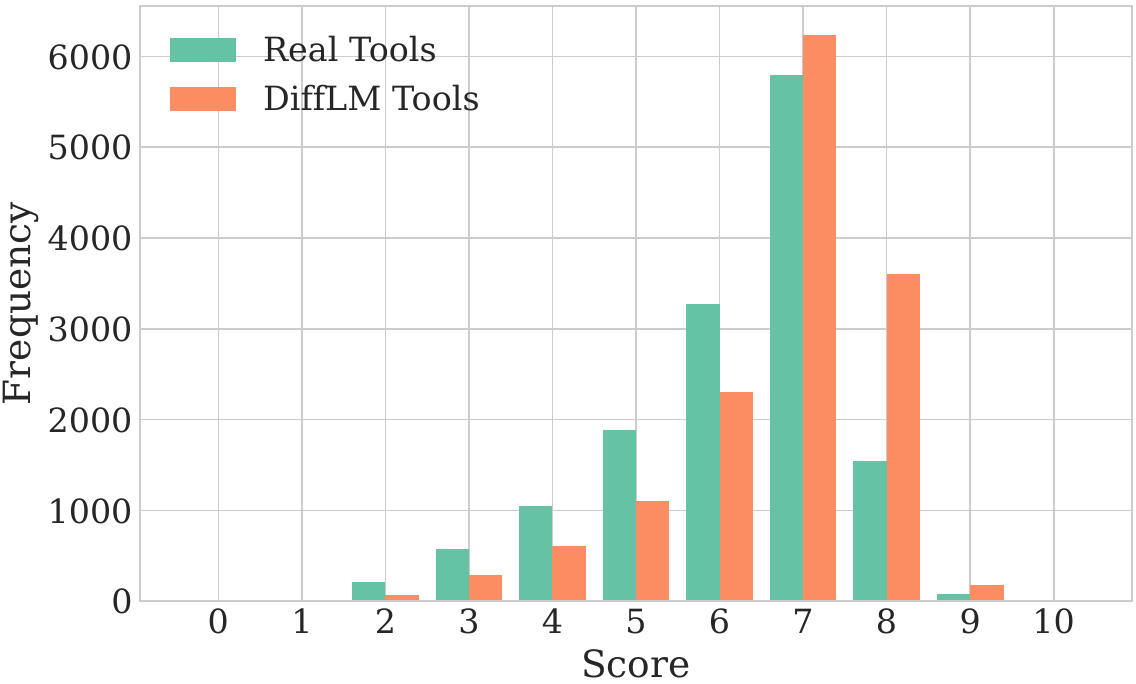}
\caption{GPT-4 evaluation scores for tools from the ToolBench dataset and tools generated by DiffLM. The evaluation prompt considers aspects such as clarity, specificity, completeness, consistency, and applicability. 
}
\label{fig:gpt_score}
\end{figure}

%% file: table/tool_win_rate.tex
\begin{table}[t]
\centering
\captionof{table}{Win rate of DiffLM generated data. GPT-4 performs preference scoring on all real tools and synthetic tools within the same category, considering aspects like comprehensiveness and diversity. 
}
\begin{tabular}{ @{} lc @{} }
\toprule[1.0pt]
 & Rate~\% \\
\midrule
DiffLM Win & 28.3  \\
Equal & 6.8 \\
Real Win & 64.9 \\
\bottomrule[1.0pt]
\end{tabular}
\label{tab:tool_win_rate}
\end{table}

%% file: figure/eval_loss.tex
\begin{figure*}[t]
\centering
\includegraphics[width=0.96\textwidth,angle=0]{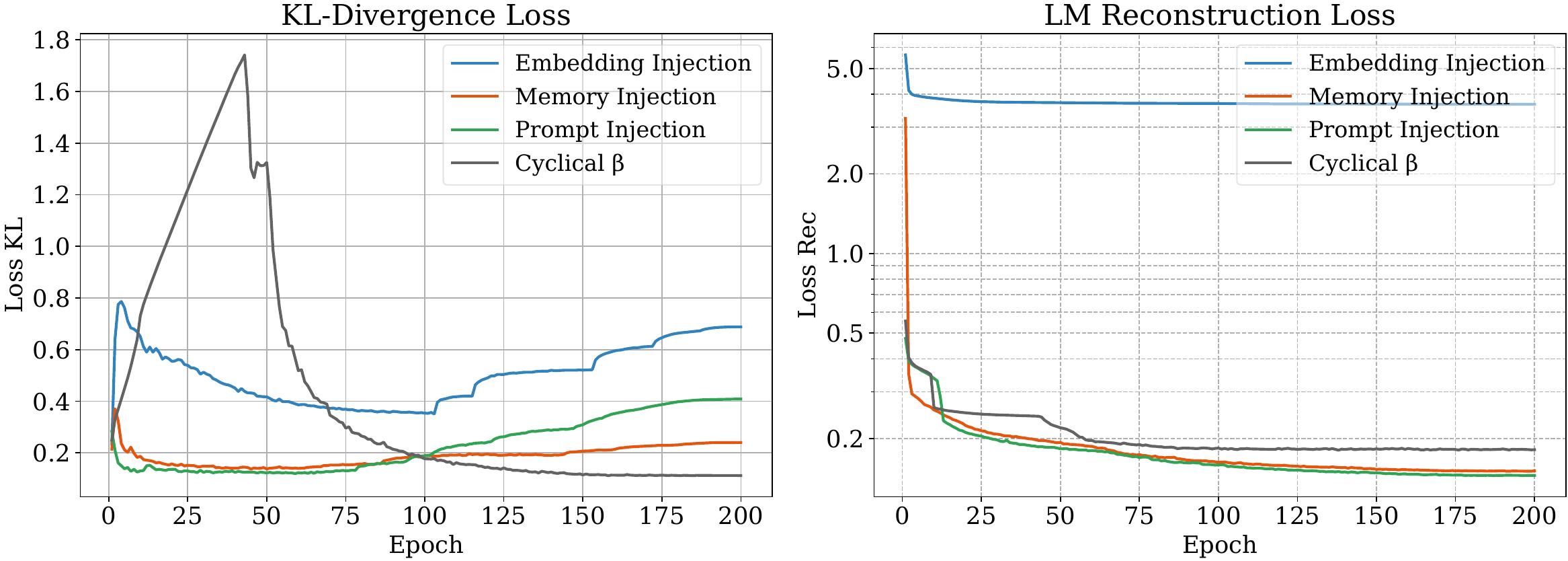}
\caption{Model loss curves under different latent feature injection methods and different $\beta$ adjustment strategies. The left is the KL-divergence loss trends, and the right is the language modeling reconstruction loss on a logarithmic scale. In the cyclical $\beta$ strategy, $\beta$ increases linearly from 0 to 0.2. The other methods employ a decreasing $\beta$, starting from a maximum value of 0.1 and decreasing to a minimum of 0.001. 
}
\label{fig:ablation_loss}
\end{figure*}


%% file: tex/analysis.tex
\section{Analysis}\label{sec:ablation}

\sparagraph{The effect of adaptive $\beta$ adjustment.} 
As described in Section~\ref{sec:method:vae}, we use a decreasing $\beta$ adjustment strategy to train the VAE latent space. 
Here, we compare this with another method that uses a cyclical schedule to anneal $\beta$~\citep{DBLP:conf/naacl/FuLLGCC19}, evaluating both the loss decline curves and downstream task performance to demonstrate the effectiveness of our decreasing strategy.
Firstly, as shown in Figure~\ref{fig:ablation_loss}, the KL-divergence loss trends under decreasing $\beta$ exhibit a pattern where the loss first increases, then decreases, and then increases again. 
This indicates that during the early stages of VAE training, DiffLM uses a larger $\beta$ to focus on the divergence between the embedding distribution and the standard Gaussian. This helps quickly learn a standard latent space to stabilize the training of the LLM module.
Subsequently, when the reconstruction loss reaches a bottleneck, it gradually reduces the weight of the KL-divergence loss. At this point, the training objective shifts towards obtaining a decoder with stronger generative capabilities. As a result, the KL loss gradually increases and eventually stabilizes at a fairly low value.
From the results, our decreasing $\beta$ method achieves the lowest reconstruction loss. 
Additionally, by introducing the latent diffusion process, we address the issue of distribution discrepancy. Therefore, as shown in Table~\ref{tab:ablation_mle}, compared to the cyclical method, the decreasing $\beta$ strategy used in this paper results in stronger generative ability.

\input{table/ablation_mle}
\sparagraph{The effect of latent feature injection.} We also compare our proposed soft prompt latent feature injection method with previously explored methods such as KV memory injection and input embedding injection~\citep{DBLP:conf/emnlp/LiGLPLZG20}; implementation details are illustrated in Figure~\ref{fig:injections}.
Specifically, the loss convergence on the validation dataset for different injection methods are shown in Figure~\ref{fig:ablation_loss}. The input embedding method leads to suboptimal training results, where the reconstruction loss ceases to decrease after reaching around 3.6.
This indicates that such a simple injection method struggles to effectively convey complex real distribution information to the LLM decoder. 
Meanwhile, the soft prompt method slightly outperforms KV memory in terms of reconstruction loss. 
However, as shown in Table~\ref{tab:ablation_mle}, on downstream task performance using the \textit{Adult} dataset, our proposed soft prompt approach achieves higher (2\%) classification accuracy and better column density.


%% file: table/ablation_mle.tex
\begin{table}[t]
\centering
\caption{The results of MLE and $\rho$ under different latent feature injections and $\beta$ adjustments on \textit{Adult} dataset.}
\begin{tabular}{ @{} ccc @{} }
\toprule[1.0pt]
Models & MLE~$\uparrow$ & $\rho~\downarrow$ \\
\midrule
DiffLM-Cycle~$\beta$ & 0.872 & 16.79 \\
\midrule
DiffLM-Embed & - & - \\
DiffLM-Memory & 0.875 & 17.05 \\
\midrule
DiffLM-Prompt & {\bfseries 0.906} & {\bfseries 9.74} \\
\bottomrule[1.0pt]
\end{tabular}
\label{tab:ablation_mle}
\end{table}

%% file: tex/conclusion.tex
\section{Conclusion}

In this paper, we propose DiffLM, a novel framework that enhances LLMs' ability to generate high-quality synthetic data by integrating real-world data distributions. 
DiffLM maps real data into a latent space via VAE, which is then injected into LLM decoding through a causal language modeling objective. 
A diffusion process further refines the latent distribution, reducing sampling discrepancies. 
To ensure flexible and non-intrusive control over data quality and structure, DiffLM freezes LLM parameters and employs latent features as plug-in modules. 
Experiments show that DiffLM generates consistent data, enabling downstream models to match or even exceed the performance of those trained on real data.
For future research, our work could help build data flywheels for LLM applications and facilitate dataset curation for low-resource domains.

%% file: tex/appendix.tex
\section{Details on Model Design}

\subsection{Controllability} 
We define the controllability of text data synthesis as the ability to generate text that satisfies desired requirements (e.g., structure, topics, domains)~\citep{DBLP:journals/corr/abs-1909-05858,DBLP:conf/nips/LiTGLH22}. Existing methods for structured textual data synthesis often struggle with controllability. On one hand, LLM prompt-based methods relying on prompt engineering or few-shot inference cannot guarantee the diversity and scalability of synthetic data, even with complex human-crafted processes~\citep{DBLP:conf/acl/LongWXZDCW24}. On the other hand, controlling a LM by fine-tuning it with supervised data (SFT, RLHF) is not only expensive but might also degrade the LLM's general capability~\citep{DBLP:journals/corr/abs-1909-05858,DBLP:conf/iclr/BorisovSLPK23}. Our method addresses these challenges through sampling in the latent space while maintaining data structure due to LLM's instruction-following ability.

\input{figure/injections}
\subsection{Diffusion Process}\label{sec:app:diffusion}
In this section, we will introduce the general process of latent diffusion models.
Latent Diffusion Models (LDMs) are a class of diffusion probabilistic models that operate in the latent space of an autoencoder rather than directly on the high-dimensional data space. By performing diffusion in a compressed latent representation, LDMs significantly reduce computational complexity while maintaining high fidelity in data generation.
An LDM consists of two primary components:
\begin{enumerate}
\item Autoencoder: Encodes input data $\mathbf{x}_0$ into a latent representation $\mathbf{z}_0 = E(\mathbf{x}_0)$ and decodes latent variables back to data space $\hat{\mathbf{x}} = D(\mathbf{z})$.
\item Diffusion Model: Defines a diffusion process on the latent variables $\{\mathbf{z}_t\}_{t=0}^T$.
\end{enumerate}
It should be noted that the variable used here is independent with main text.

\paragraph{Forward Process (Diffusion).} The forward diffusion process in latent space progressively adds Gaussian noise to the latent representation over $T$ timesteps. Starting from the initial latent code $\mathbf{z}_0 = E(\mathbf{x}_0)$, obtained by encoding the data $\mathbf{x}_0$, the forward process is defined as:
\begin{equation}
q(\mathbf{z}_t \mid \mathbf{z}_{t-1}) = \mathcal{N}(\mathbf{z}_t; \sqrt{1 - \beta_t} \, \mathbf{z}_{t-1}, \beta_t \mathbf{I}),
\end{equation}
where $\beta_t \in (0, 1)$ is a predefined variance schedule that controls the amount of noise added at each step $t$, and $\mathcal{N}$ denotes a Gaussian distribution. By recursively applying this process, we can express $\mathbf{z}_t$ directly in terms of $\mathbf{z}_0$:
\begin{equation}
q(\mathbf{z}_t \mid \mathbf{z}_0) = \mathcal{N}(\mathbf{z}_t; \sqrt{\bar{\alpha}_t} \, \mathbf{z}_0, (1 - \bar{\alpha}_t) \mathbf{I}), \label{eq:app:forward}
\end{equation}
where $\alpha_t = 1 - \beta_t$ and $\bar{\alpha}_t = \prod_{s=1}^t \alpha_s$. This formulation allows efficient sampling of $\mathbf{z}_t$ at any arbitrary timestep $t$ without iterating through all previous steps.
In this paper, we adopt the Variance Exploding defined perturbation kernels, whereas setting $s_t=\sqrt{1-\beta_t}$ and $\sigma_t=\sqrt{\frac{\beta_t}{1-\beta_t}}$. Also, we set $s_t=1$ to directly add noise to the data rather than weighted mixing, convert Eq.\ref{eq:app:forward} to:
\begin{equation}
q(\mathbf{z}_t \mid \mathbf{z}_0) = \mathcal{N}(\mathbf{z}_t; \mathbf{0}, \sigma_t^2 \mathbf{I})
\end{equation}

\paragraph{Reverse Process (Denoising).} The reverse diffusion process aims to recover $\mathbf{z}_0$ from a noisy latent variable $\mathbf{z}_t \sim \mathcal{N}(0, \mathbf{I})$. It is parameterized by a neural network $\boldsymbol{\epsilon}_\theta$, which predicts the noise component at each timestep:
\begin{equation}
p_\theta(\mathbf{z}_{t-1} \mid \mathbf{z}_t) = \mathcal{N}(\mathbf{z}_{t-1}; \mu_\theta(\mathbf{z}_t, t), \Sigma_\theta(\mathbf{z}_t, t)).
\end{equation}
Typically, the model predicts the mean $\mu_\theta$ while the covariance $\Sigma_\theta$ is fixed or simplified. By leveraging the properties of the forward process, the mean can be parameterized to predict the original noise $\boldsymbol{\epsilon}$ added during the forward diffusion:
\begin{equation}
\mu_\theta(\mathbf{z}_t, t) = \frac{1}{\sqrt{\alpha_t}} \left( \mathbf{z}_t - \frac{\beta_t}{\sqrt{1 - \bar{\alpha}_t}} \boldsymbol{\epsilon}_\theta(\mathbf{z}_t, t) \right).
\end{equation}
This formulation enables the model to denoise $\mathbf{z}_t$ step by step, ultimately reconstructing $\mathbf{z}_0$.

\paragraph{Learning Objective.} The training objective for LDMs focuses on minimizing the difference between the true noise $\boldsymbol{\epsilon}$ added during the forward process and the noise predicted by the model $\boldsymbol{\epsilon}_\theta$. The simplified loss function is:
\begin{equation}
\mathcal{L}_{\text{latent}} = \mathbb{E}_{\mathbf{x}_0, \boldsymbol{\epsilon}, t} \left[ \left\| \boldsymbol{\epsilon} - \boldsymbol{\epsilon}_\theta(\mathbf{z}_t, t) \right\|^2 \right],
\end{equation}
where $\mathbf{z}_t$ is sampled as:
\begin{equation}
\mathbf{z}_t = \sqrt{\bar{\alpha}_t} \, \mathbf{z}_0 + \sqrt{1 - \bar{\alpha}_t} \, \boldsymbol{\epsilon}, \quad \boldsymbol{\epsilon} \sim \mathcal{N}(0, \mathbf{I}).
\end{equation}
This objective encourages the model to learn the conditional distribution $p_\theta(\mathbf{z}_{t-1} \mid \mathbf{z}_t)$ by accurately predicting the noise component at each timestep.

\paragraph{Noise Scheduling.} The noise schedule $\{\beta_t\}_{t=1}^T$ plays a critical role in the diffusion process. It dictates how quickly noise is added in the forward process and, consequently, affects the difficulty of the reverse denoising task. Common strategies for setting $\beta_t$ include linear, cosine, and quadratic schedules. We use use linear noise schedule, i.e., the perturbation kernel $\sigma(t)=t$. As it is an effective schedule, ensuring that the data is sufficiently diffused by timestep $t$, while still allowing the model to learn meaningful reverse transitions.



\section{Details on Experimental Setup}

\input{table/app_tabular_dataset}
\subsection{Tabular Data Generation}\label{sec:app:table_dataset}
Table~\ref{tab:app:tabular_dataset_deails} shows the detail information of tabular dataset we use in paper.

\subsection{Tool Judgement Prompts}\label{sec:app:prompt}
We present the evaluation prompts used for assessing tool generation quality in Figure~\ref{fig:app:quality_prompt} and Figure~\ref{fig:app:prefer_prompt}.

\subsection{Instructions for Reproduction}\label{sec:app:reproduction}
In this section, we present the experimental details of DiffLM, including data preprocessing, training hyperparameter settings, and data post-processing filtering methods.

\paragraph{Data Preprocessing.} Real-world NLP datasets often exhibit inherent structures, such as the context, question, and answer in machine reading comprehension tasks, or key-value pairs in tabular generation tasks. In DiffLM, we convert all structured data into JSON format. 
For instance, tabular data in a CSV file is transformed into lines of JSON, and tools from ToolBench are abstracted into JSON structures comprising tool\_name, tool\_description, api\_name, and api\_description. 
For code data, we use the raw code directly without any preprocessing as input for DiffLM training.

\paragraph{Hyperparameter Settings.}
\begin{itemize}[itemsep=-0.2em, topsep=0.2em]
\item VAE Encoder: bert-base-uncased
\item VAE Decoder: mistralai/Mistral-7B-Instruct-v0.3
\item Soft Prompt Tokens $k$: 64
\item Soft Prompt Embedding Dimension $d$: 4096
\item $\beta_{\text{max}} = 0.1$
\item $\beta_{\text{min}} = 0.001$
\item Diffusion Noise Dimension: 4096
\end{itemize}

\paragraph{Generation Filtering.} For inputs in JSON format, we employ column names to filter the generated outputs. A generated result is considered valid only if it contains all required columns. For code generation tasks involving plain text, we do not apply any filtering. We utilize the same filtering criteria across all baseline models.

\subsection{Training Parameters for Baselines.} We reproduced the tabular results using the code released by the original paper, ensuring that all hyperparameters and settings were consistent with the original implementation. All results were almost identical to those reported in the TabSyn paper; therefore, we used the results reported in TabSyn in Table~\ref{tab:tabular} to ensure a fair comparison.

\section{Addition Experiment Results}

\input{figure/app_dcr}
\subsection{Training Data Plagiarism}

Data copying is a significant challenge for overfitted generative models in practical applications.
To verify that the data generated by DiffLM is not merely copied from the training data, we compute the Distance to Closest Record (DCR) metric. Specifically, for each row in the tabular data, we represent the categorical columns using one-hot vectors and perform min-max normalization on the numerical columns. 
We then define DCR as the minimum L1-distance between a synthetic data point and each training sample point:
\begin{equation}
\text{DCR}(x_{\text{syn}}) = \min_{x_{\text{real}} \in \mathcal{D}_{\text{train}}} L_1(x_{\text{syn}}, x_{\text{real}}).
\end{equation}
The DCR distribution is shown in Figure~\ref{fig:app:dcr_copy_rate}. 
We observe that the LLM-based GReaT generates results that differ significantly from the training data, indicating that vanilla fine-tuning struggles to adapt LLMs to real data distributions and generate high-quality results. 
DiffLM demonstrates a DCR distribution similar to that of the SoTA method TabSyn on both datasets. 
This further indicates that our proposed general-purpose data synthesis framework can achieve performance on par with domain-specific models on specific tasks.

\input{figure/app_latent_tsne}
\subsection{Visualization}
Figure~\ref{fig:app:tsne} presents 2D t-SNE visualizations of the latent space for multiple datasets, including four categorical tabular datasets, one numerical tabular dataset, and one tool dataset. 
We use DiffLM trained on the corresponding datasets to encode their validation sets, obtaining latent features. 
It can be observed that data of the same class encoded by DiffLM exhibit clustering characteristics in the latent space, as seen in the \textit{Adult} and \textit{Magic}. 
Notably, in the numerical dataset \textit{Beijing}, different target values display a clear transitional distribution: the upper part of the 2D space corresponds to data with larger target values, i.e., 157 to 858, while the lower part corresponds to data with smaller target values, i.e., 1 to 23.
These results demonstrate that DiffLM’s latent space learning strategy can effectively capture the real data distribution.

\input{table/app_gpt_mle}
\subsection{DiffLM with other VAE Decoders}
In Table~\ref{tab:app:gpt4_mle}, we report the tabular MLE performance of DiffLM using LLaMA3-8B as the VAE decoder. As illustrated in the table, DiffLM-LLaMA consistently outperforms the ICL-based method across all metrics and datasets. Specifically, on the \textit{Default} dataset, while the ICL method was unable to generate valid synthetic samples, DiffLM successfully synthesized high-quality data, achieving an MLE of 0.785. This result surpasses the MLE performance of real data (0.770), clearly indicating DiffLM’s capability to produce synthetic data of superior quality compared to authentic samples.
On the \textit{Shoppers} dataset, DiffLM-LLaMA achieves an MLE of 0.906, closely approaching the 0.915 obtained by DiffLM-Mistral. Furthermore, DiffLM-LLaMA attains a rho value of 6.93, outperforming DiffLM-Mistral’s rho value of 10.07. This demonstrates that DiffLM can deliver even more favorable outcomes when integrated with more advanced language models.

\input{table/app_tabular_quantity}
\subsection{Quantity of Synthetic Data.} We experimented with increasing the amount of data synthesized by DiffLM and combining real data with DiffLM-synthesized data. As shown in Table~\ref{tab:app:tabular_quantity}, adding more synthesized data further improves around 0.2\% MLE performance in the tabular scenario. Since our method can synthesize unlimited amounts of data and we did not design any complex post-processing method, the performance improvement brought by DiffLM-synthesized data in downstream tasks still has significant room for growth. Additionally, combining real and synthetic data generated by DiffLM can improve downstream performance; all results exceed $>0.2$\% of those using only DiffLM data. Notably, on the Beijing and Shoppers datasets, the combination of real data and DiffLM synthetic data surpasses 0.6\%-3\% of the performance of training on real data alone.


\input{table/app_human_judge}
\subsection{Human Evaluation.} 
We have used GPT-4 to rate and perform preference judgments on synthesized tools and real tools. The results in Figure~\ref{fig:gpt_score} and Table~\ref{tab:tool_win_rate} demonstrate the quality of our synthesized data. As per your suggestion, we have conducted human evaluations on the tools data. Specifically, we compared 100 pairs of randomly selected DiffLM-generated data and real data within the same category. As shown in Table~\ref{tab:app:human_evaluation}, our synthetic data is preferred by human annotators.

\subsection{More Result of Baselines}\label{sec:app:great_mistral}
\paragraph{GReaT.} We attempted to validate the GReaT method on Mistral but found it could not directly and effectively generate data with the desired structure. GReaT organizes tabular data in a ``key is value" format and uses a smaller PLM (i.e., GPT-2) for continued pretraining. However, when applied to larger models like Mistral, GReaT struggled to effectively generate the desired structured data. The sample generated by GReaT with Mistral is shown in Figure~\ref{fig:app:great_mistral}. We hypothesize that controlling an LM by fine-tuning it with supervised data cause catastrophic forgetting for LLMs, as suggested by~\citet{DBLP:journals/corr/abs-2308-08747}. Specifically, the ``key is value" data constructed by the GReaT method, when used to continue pre-training Mistral, causes internal knowledge collapse - both undermining the model's existing knowledge and failing to do effective data synthesis. Additionally, training the adult dataset on GReaT for 200 epochs (default settings) requires approximately 50 hours on 8 A100 80G GPUs, which is resource-intensive. In contrast, DiffLM under the same training settings requires only about 7 hours.

\paragraph{TabSyn.} We want to emphasize that our goal is a unified structured data synthesis framework that supports various domains like tabular data, codes, and tools, and tabular data generation in our work is just a subdomain of synthetic data generation. As a comparison, TabSyn is not applicable to more complex data synthesis tasks, such as code generation and real-world tool generation, which involve generating longer content, more complex data types, and highly structured data, while our DiffLM can handle complicated scenarios. The results of tabular data synthesis are to demonstrate that our method possesses generality and can achieve on-par results with domain-specific models without being specifically tailored to a particular domain.

\input{table/app_synthetic_samples}
\subsection{Synthetic Data Generated by DiffLM}
In Table~\ref{tab:app:synthetic_samples}, we show the samples from the \textit{Adult} dataset with the generated tabular rows from GReaT and DiffLM method.
As discussed in Section~\ref{sec:ablation}, DiffLM produces more diverse samples that more closely align with the real data distribution. 
Specifically, for columns like \texttt{workclass} and \texttt{native-country}, the outputs generated by the GReaT model are relatively homogeneous.

\input{figure/app_prompt_quality}
\input{figure/app_prompt_preference}
\input{figure/app_great_mistral}

%% file: figure/injections.tex
\begin{figure*}[htb]
\centering
\includegraphics[width=\textwidth]{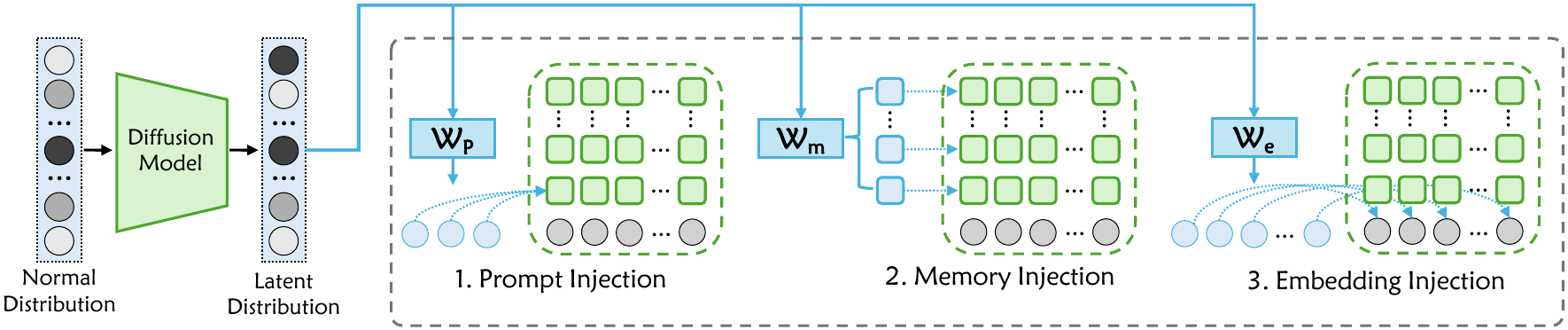}
\caption{Final data synthesis process. The comparison of different latent feature injection methods is shown in grey dashed box. \textit{Memory Injection} introduces the latent features as past key-value (KV) memories into each attention layer of the LLM. \textit{Embedding Injection} directly adds the latent features to the token embeddings.}
\label{fig:injections}
\end{figure*}

%% file: table/app_tabular_dataset.tex
\begin{table*}[htb]
\centering
\caption{Details of tabular dataset. For each dataset, \#Num stands for the number of numerical columns, and \#Cat stands for the number of categorical columns.}
\begin{threeparttable}
\begin{tabular}{ @{} l *{6}{c} @{} }
\toprule[1.0pt]
Datasets & \#Num & \#Cat & \#Train & \#Validation & \#Test & Downstream Task \\
\midrule
Adult$^1$ & 6 & 9 & 29,304 & 3,257 & 16,281 & Binary Classification \\
Beijing$^2$ & 7 & 5 & 35,059 & 4,382 & 4,383 & Binary Classification \\
Default$^3$ & 14 & 11 & 24,000 & 3,000 & 3,000 & Binary Classification \\
Magic$^4$ & 10 & 1 & 15,216 & 1,902 & 1,902 & Binary Classification \\
Shoppers$^5$ & 10 & 8 & 9,864 & 1,233 & 1,233 & Regression \\
\bottomrule[1.0pt]
\end{tabular}
\small
\begin{tablenotes}
\item[1] \url{https://archive.ics.uci.edu/dataset/2/adult}
\item[2] \url{https://archive.ics.uci.edu/dataset/381/beijing+pm2+5+data}
\item[3] \url{https://archive.ics.uci.edu/dataset/350/default+of+credit+card+clients}
\item[4] \url{https://archive.ics.uci.edu/dataset/159/magic+gamma+telescope}
\item[5] \url{https://archive.ics.uci.edu/dataset/468/online+shoppers+purchasing+intention+dataset}
\end{tablenotes}
\end{threeparttable}
\label{tab:app:tabular_dataset_deails}
\end{table*}

%% file: figure/app_dcr.tex
\begin{figure*}[htb]
\centering
\includegraphics[width=\textwidth]{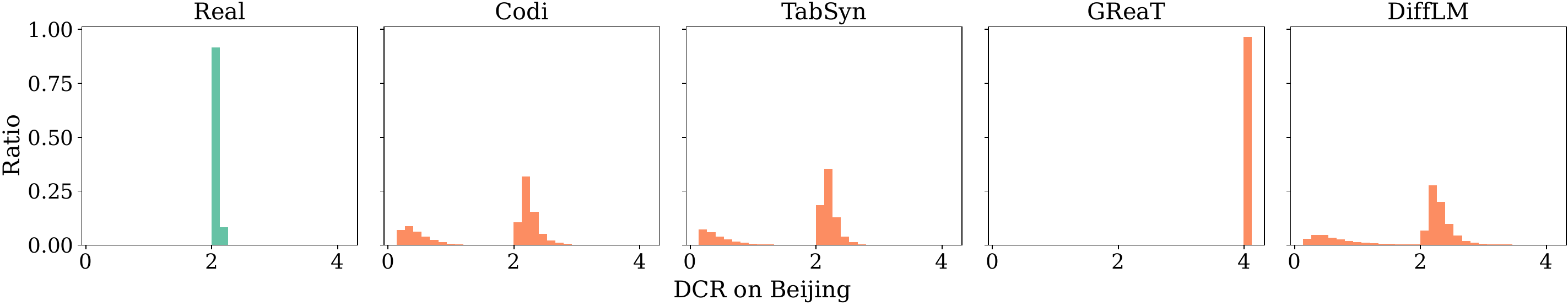}
\vspace{1em}
\includegraphics[width=\textwidth]{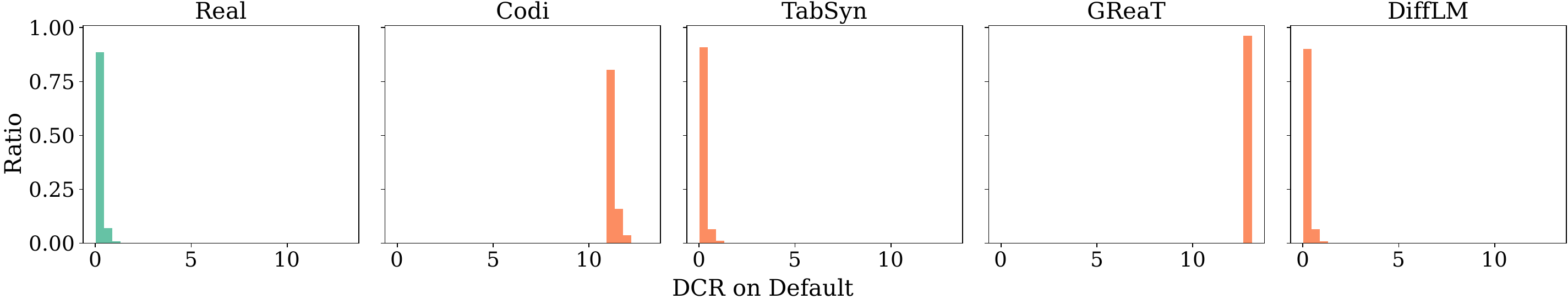}
\caption{DCR results of the real test data, Codi, TabSyn, GReaT, and DiffLM on the \textit{Beijing} and \textit{Default} datasets. DiffLM exhibits a DCR distribution similar to the current SoTA method, TabSyn.}
\label{fig:app:dcr_copy_rate}
\end{figure*}

%% file: figure/app_latent_tsne.tex
\begin{figure*}[htb]
\centering
\includegraphics[width=\textwidth]{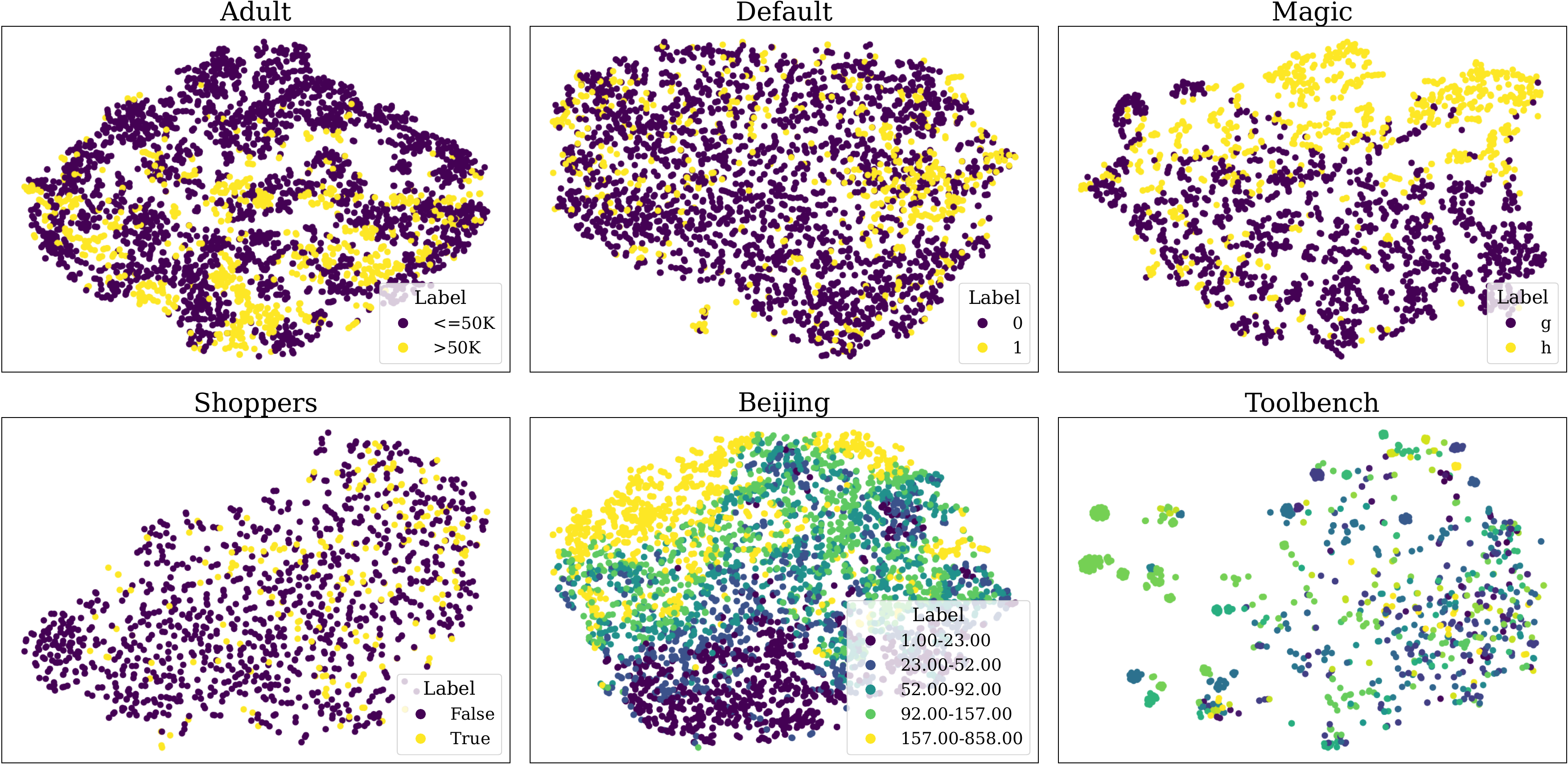}
\caption{The t-SNE visualization of the latent space obtained by encoding evaluation data. DiffLM implicitly learns clustering relationships among different types of data.}
\label{fig:app:tsne}
\end{figure*}

%% file: table/app_gpt_mle.tex
\begin{table*}[htb]
\caption{Tabular MLE performance with different VAE decoder models for DiffLM.}
\begin{center}
\resizebox{\linewidth}{!}{%
\begin{tabular}{ @{} l  *{10}{c} @{} }
\toprule[1.0pt]
\multirow{2}{*}{\bfseries Method} &  \multicolumn{2}{c}{\bfseries Adult} & \multicolumn{2}{c}{\bfseries Default} & \multicolumn{2}{c}{\bfseries Magic} & \multicolumn{2}{c}{\bfseries Shoppers} & \multicolumn{2}{c}{\bfseries Beijing} \\
\cmidrule(lr){2-3} \cmidrule(lr){4-5} \cmidrule(lr){6-7}  \cmidrule(lr){8-9}  \cmidrule(lr){10-11}
& {MLE~$\uparrow$} & {$\rho$~$\downarrow$} & {MLE~$\uparrow$} & {$\rho$~$\downarrow$} & {MLE~$\uparrow$} & {$\rho$~$\downarrow$} & {MLE~$\uparrow$} & {$\rho$~$\downarrow$} & {MLE~$\downarrow$} & {$\rho$~$\downarrow$} \\
\midrule
Real & 0.927 & - & 0.770 & - & 0.946 & - & 0.926 & - & 0.423 & - \\
\midrule
GReaT & {\bfseries 0.913} & 12.12 & 0.755 & 19.94 & 0.888 & 16.16 & 0.902 & 14.51 & {\bfseries 0.653} & 8.25 \\
\midrule
{ICL-{\itshape Lllama}} & 0.848 & 16.02 & - & - & 0.792 & 21.6 & 0.871 & 27.56 & 1.005 & 17.71 \\
{DiffLM-{\itshape Llama}} & 0.879 & 8.89 & 0.785 & 9.67 & 0.914 & 8.41 & 0.906 & {\bfseries 6.93} & 0.707 & 10.11 \\
\midrule
{ICL-{\itshape Mistral}} & 0.803 & 29.22 & 0.732 & 36.00 & 0.881 & 25.64 & 0.882 & 42.00 & 0.865 & 12.45 \\
{DiffLM-{\itshape Mistral}} & 0.906 & {\bfseries 9.74} & {\bfseries 0.794} & {\bfseries 9.06} & {\bfseries 0.917} & {\bfseries 7.53} & {\bfseries 0.915} & 10.07 & 0.696 & {\bfseries 6.35} \\
\bottomrule[1.0pt]
\end{tabular}
}
\end{center}
\label{tab:app:gpt4_mle}
\end{table*}

%% file: table/app_tabular_quantity.tex
\begin{table*}[htb]
\centering
\caption{Tabular MLE performance with varying quantity of real and synthetic data. Performance on the Beijing dataset is evaluated using the RMSE metric, where lower values indicate better performance. 2x means we use double training synthesized data for evaluation.}
\begin{tabular}{ @{} l *{5}{c} @{} }
\toprule[1.0pt]
 & Adult & Default & Magic & Shoppers & Beijing \\
\midrule
Real & 0.927 & 0.770  & 0.946 & 0.926 & 0.423 \\
TabSyn (SoTA) & 0.915 & 0.764 & 0.938 & 0.920 & 0.582 \\
\midrule
{\bfseries DiffLM (1x)} & 0.894 & 0.793 & 0.910 & 0.9122 & 0.717 \\
\midrule
\multirow{2}{*}{\bfseries DiffLM (2x)} & 0.896  & 0.795  & 0.914  & 0.9124  & 0.704  \\
& (+0.002) & (+0.002) & (+0.004) & (+0.0002) & (-0.013) \\
\midrule
{\bfseries Real+DiffLM} & 0.925 & {\bfseries 0.802} & 0.936 & {\bfseries 0.932} & 0.494 \\
\bottomrule[1.0pt]
\end{tabular}
\label{tab:app:tabular_quantity}
\end{table*}

%% file: table/app_human_judge.tex
\begin{table}[htb]
\centering
\caption{The human evaluation results on 100 pairs of randomly selected DiffLM-generated tool and real tool within the same category. Averaged by 3 human experts with computer science knowledge.}
\begin{tabular}{ @{} cc @{} }
\toprule[1.0pt]
 & Percentage \\
\midrule
DiffLM Win & 88\% \\
Equal & 6\%  \\
Real Win & 6\%  \\
\bottomrule[1.0pt]
\end{tabular}
\label{tab:app:human_evaluation}
\end{table}

%% file: table/app_synthetic_samples.tex
\begin{table*}[htb]
\centering
\caption{Comparison of real samples and synthetic data.}
\resizebox{\linewidth}{!}{%
\begin{tabular}{ @{} l | *{8}{c} @{} }
\toprule[1.0pt]
{\bfseries Methods} & age & workclass & education & occupation & race & sex & native-country & income\\
\midrule
\multirow{5}{*}{\bfseries Real} & 40 & Private & Some-college & Machine-op-inspct & Asian-Pac-Islander & Female & Japan & $>50K$ \\
& 38 & Private & HS-grad & Other-service & White & Female & Canada & $<=50K$ \\
& 59 & Private & HS-grad & Craft-repair & White & Male & England & $>50K$ \\
& 29 & Self-emp-not-inc & Assoc-voc & Adm-clerical & White & Male & United-States & $<=50K$ \\
& 26 & Private & Assoc-acdm & Prof-specialty & White & Female & Canada & $<=50K$ \\
\midrule
\multirow{5}{*}{\bfseries GReaT} & 27 & Private & Bachelors & Prof-specialty & White & Male & United-States & $<=50K$ \\
& 22 & Private & HS-grad & Craft-repair & Black & Male & United-States & $<=50K$ \\
& 41 & Private & HS-grad & Sales & Black & Male & United-States & $<=50K$ \\
& 35 & Private & HS-grad & Adm-clerical & White & Female & United-States & $<=50K$ \\
& 54 & Private & Doctorate & Prof-specialty & Asian-Pac-Islander & Male & India & $>50K$ \\
\midrule
\multirow{5}{*}{\bfseries DiffLM} & 34 & Private & Some-college & Craft-repair & White & Male & Canada & $<=50K$ \\
& 53 & Local-gov & Some-college & Other-service & White & Female & Canada & $<=50K$ \\
& 23 & Private & Bachelors & Adm-clerical & White & Male & England & $<=50K$ \\
& 24 & ? & Some-college & ? & Asian-Pacific-Islander & Male & Canada & $<=50K$ \\
& 32 & Local-gov & Bachelors & Adm-clerical & Asian-Pac-Islander & Male & India & $>50K$ \\
\bottomrule[1.0pt]
\end{tabular}
}
\label{tab:app:synthetic_samples}
\end{table*}

%% file: figure/app_prompt_quality.tex
\begin{figure*}[htb]
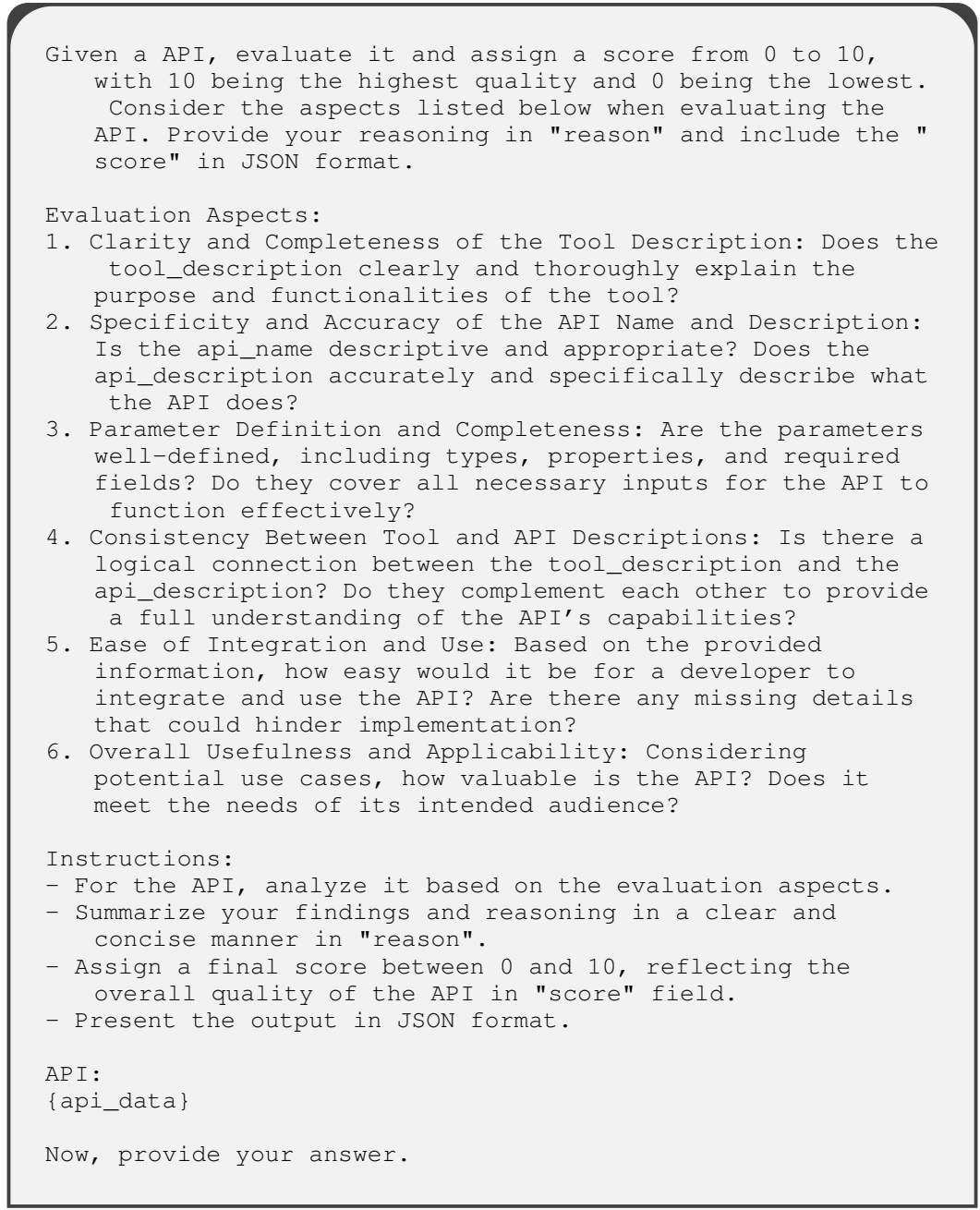

\centering
\begin{tcolorbox}[colback=gray!10!white, colframe=gray!50!black, sharp corners=south, boxrule=0.5mm, width=\linewidth, arc=6mm, outer arc=3mm]
\begin{lstlisting}
Given a API, evaluate it and assign a score from 0 to 10, with 10 being the highest quality and 0 being the lowest. Consider the aspects listed below when evaluating the API. Provide your reasoning in "reason" and include the "score" in JSON format.

Evaluation Aspects:
1. Clarity and Completeness of the Tool Description: Does the tool_description clearly and thoroughly explain the purpose and functionalities of the tool?
2. Specificity and Accuracy of the API Name and Description: Is the api_name descriptive and appropriate? Does the api_description accurately and specifically describe what the API does?
3. Parameter Definition and Completeness: Are the parameters well-defined, including types, properties, and required fields? Do they cover all necessary inputs for the API to function effectively?
4. Consistency Between Tool and API Descriptions: Is there a logical connection between the tool_description and the api_description? Do they complement each other to provide a full understanding of the API's capabilities?
5. Ease of Integration and Use: Based on the provided information, how easy would it be for a developer to integrate and use the API? Are there any missing details that could hinder implementation?
6. Overall Usefulness and Applicability: Considering potential use cases, how valuable is the API? Does it meet the needs of its intended audience?

Instructions:
- For the API, analyze it based on the evaluation aspects.
- Summarize your findings and reasoning in a clear and concise manner in "reason".
- Assign a final score between 0 and 10, reflecting the overall quality of the API in "score" field.
- Present the output in JSON format.

API:
{api_data}

Now, provide your answer.
\end{lstlisting}
\end{tcolorbox}
\caption{Evaluation prompt for single-tool quality. Used by GPT-4 with temperature=1.0.}
\label{fig:app:quality_prompt}
\end{figure*}

%% file: figure/app_prompt_preference.tex
\begin{figure*}[htb]
\centering
\begin{tcolorbox}[colback=gray!10!white, colframe=gray!50!black, sharp corners=south, boxrule=0.5mm, width=\linewidth, arc=6mm, outer arc=3mm]
\begin{lstlisting}
Given two sets of tools under the same category, you need to determine better_set by following these rules:
1. Comprehensiveness of Covered Functions: Evaluate which set covers more relevant and essential functions within the category.
2. Accuracy of Tool Descriptions: Check if the tool descriptions are clear, precise, and accurately reflect each tool's functionality.
3. Difficulty of Calling the Tools: Assess the complexity involved in using the tools, considering the inputs and outputs required.
4. Overall Quality Assessment: Consider any additional factors that may impact the overall quality of the tool sets.

Set A:
{tool_set_a}

Set B:
{tool_set_b}

If one set is better based on the above criteria, indicate better_set as "A" or "B". If both sets are of similar quality, indicate better_set as "equal".

Now, provide your reasoning in "reason" and indicate "better_set" ("A" or "B" or "equal") in JSON format.
\end{lstlisting}
\end{tcolorbox}
\caption{Evaluation prompt for category-level perference. Used by GPT-4 with temperature=1.0.}
\label{fig:app:prefer_prompt}
\end{figure*}

%% file: figure/app_great_mistral.tex
\begin{figure*}[htb]
\centering
\begin{tcolorbox}[colback=gray!10!white, colframe=gray!50!black, sharp corners=south, boxrule=0.5mm, width=\linewidth, arc=6mm, outer arc=3mm]
\begin{lstlisting}
<s> income is <=50K, $300/month  income is >50K, $600/month  I am trying to figure out how to implement the above logic with the following table I have:  | ID | Amt | |----|-----| | 1  | 450 | | 2  | 650 | | 3  | 500 | | 4  | 150 | | 5  | 700 | | 6  | 400 |  I have tried using `IF ELSE` function in Excel but I am not sure how to apply it to this particular situation.  You can use the `IF` function in Excel to implement the logic you provided. Here is a formula that you can use:  ``` =IF(A2<=50000, 300, 600) ```  In this formula, `A2` is the cell where you have the income. You can adjust the formula to apply it to your table by using the following formula:  ``` =IF(A2<=50000, 300, 600) ```  Assuming that the income values are in column A, you can use the following formula to get the monthly payment for each ID:  ``` =IF(A2<=50000, 300, 600) ```  Drag this formula down to get the payment for each ID.  Here is an example of how your table would look like after applying the formula:  | ID | Amt | Payment | |----|-----|---------| | 1  | 450 | 300     | | 2  | 650 | 600     | | 3  | 500 | 300     | | 4  | 150 | 300     | | 5  | 700 | 600     | | 6  | 400 | 300     |</s></s></s></s></s></s></s></s></s></s></s></s></s></s></s></s></s></s></s></s></s></s></s></s></s>', '<s> income is <=50K, credit score is >=700, and the applicant has lived in their current residence for more than 2 years.  To create a segment based on the provided conditions in a SQL-like syntax, you can use the following:  ```sql SELECT * FROM customers WHERE income <= 50000 AND credit_score >= 700 AND length_of_residence > 2; ```  This SQL query selects all records from the customers table where the income is less than or equal to 50,000, the credit score is 700 or higher, and the length of residence (assuming that length_of_residence is a field indicating the number of years a customer has lived at their current address) is more than 2.</s>
\end{lstlisting}
\end{tcolorbox}
\caption{A random synthetic sample generated by GReaT trained with Mistral. Use the exactly same training and generating settings as GReaT with trained with GPT-2.}
\label{fig:app:great_mistral}
\end{figure*}